\documentclass{article}
\usepackage{spconf,amsmath,epsfig}
\usepackage{comment}
\usepackage{multirow}
\graphicspath{{./figures/}}

\usepackage[
  acronym,
]{glossaries-extra}

\setabbreviationstyle[acronym]{long-short}
\defglsentryfmt{\ifglsused{\glslabel}{\glsgenentryfmt}{\textit{\glsgenentryfmt}}}
\defglsentryfmt[acronym]{\ifglsused{\glslabel}{\glsgenentryfmt}{\textit{\glsgenentryfmt}}}
\glsdisablehyper

\newcommand{\class}[1]{\texttt{#1}}
\newacronym{gl:lrp}{LRP}{layer-wise relevance propagation}
\newacronym{gl:xai}{XAI}{eXplainable AI}
\newacronym{gl:dl}{DL}{deep Learning}
\newacronym{gl:mlp}{MLP}{multi-layer perceptron}

\newacronym{gl:sunflower}{\class{Sunflower}}{\class{Sunflower and Yellow Bloomers}}
\newacronym{gl:corn}{\class{Maize}}{\class{Grain Maize}}

\usepackage{xargs}

\usepackage{siunitx}
\usepackage[UKenglish]{babel}
\usepackage[useregional]{datetime2}

\usepackage{xspace}
\usepackage{microtype}

\usepackage{interval}

\newcommand{\relevance}{\ensuremath{R}\xspace}

\newcommandx*{\timerange}[4][1=,4={\,--\,}]{\DTMlangsetup{showyear=false,ord=raise}\DTMlangsetup{#1}\DTMDate{#2}#4\DTMDate{#3}}

\usepackage[pdftex=true,breaklinks=true,hidelinks=true,colorlinks=true,citecolor=blue]{hyperref}
\makeatletter
\define@key{href}{font}{#1}
\makeatother
\usepackage{xpatch}
\newcommand\hrefdefaultfont{\ttfamily\small}
\xpatchcmd\href{\setkeys{href}{#1}}{\setkeys{href}{font=\hrefdefaultfont,#1}}{}{\fail}

\usepackage[
  backend=biber,
  citestyle=numeric-comp,
  bibstyle=ieee,
  sorting=nyt,
  backref=false,
  dashed=false,
  alldates=year,
  maxcitenames=2,
  mincitenames=1,
  url=false,
  doi=true,
  eprint=false,
]{biblatex}

\DeclareNameAlias{sortname}{last-first}
\AtBeginBibliography{\ninept}
\DeclareSourcemap{
  \maps[datatype=bibtex, overwrite]{
    \map{
      \step[fieldset=editor, null]
    }
  }
}

\usepackage[
  capitalize,
]{cleveref}

\usepackage{tikz,pgf}
\usetikzlibrary{positioning}
\usepackage{tumcolors}

\usepackage{subcaption}

\title{XAI for Early Crop Classification}
\name{%
  Ayshah Chan%
  \textsuperscript{1,2,*},
	Maja Schneider%
  \textsuperscript{1}, and
	Marco K{\"o}rner%
  \textsuperscript{1,2},
  \thanks{%
    \textsuperscript{*}corresponding author\newline\footnotesize
    M. Schneider, A. Chan, and the EuroCrops project receive funding from the German \emph{Federal Ministry for Economics and Climate Action} on the basis of a decision by the German Bundestag under funding reference \texttt{50EE2105}.
    The authors acknowledge support by the Munich Data Science Institute (MDSI) at Technical University of Munich (TUM) via the Linde/MDSI PhD Fellowship program and from the \emph{Stifterverband für die Deutsche Wissenschaft} for supporting the EuroCrops project with the Open Data Impact Award 2021.
  }
}

\address{%
  \parbox{\linewidth}{\centering%
    \textsuperscript{1}~%
    Technical University of Munich (TUM),
    TUM School of Engineering and Design,\\
    Chair of Remote Sensing Technology,
    Arcisstr. 21, Munich, Germany\\
  }\\
  \parbox{\linewidth}{\centering%
    \textsuperscript{2}~%
    Munich Data Science Institute (MDSI)
  }\\
  \href{mailto:ayshah.chan@tum.de}{\{ayshah.chan},
  \href{mailto:maja.schneider@tum.de}{maja.schneider},
  \href{mailto:marco.koerner@tum.de}{marco.koerner\}}%
  \href{mailto:ayshah.chan@tum.de}{@tum.de}
}
\begin{document}
\maketitle
\makeatletter
\newrobustcmd*{\setmaxcitenames}{\numdef\blx@maxcitenames}
\makeatother
\begin{refsection}
  \setmaxcitenames{3}%
  \tikz[remember picture, overlay] {%
    \node[draw=tumivory, anchor=north, below=of current page.north, text width=.825\paperwidth, inner sep=.5em, font=\small] {%
      This paper 
      will be
      published as:\newline
      \fullcite*{Chan23:XAI}
    };
  }%
\end{refsection}
\begin{abstract}
  We propose an approach for early crop classification through identifying important timesteps with \gls{gl:xai} methods.
  Our approach consists of training a baseline crop classification model to carry out \gls{gl:lrp} so that the salient time step can be identified.
  We chose a selected number of such important time indices to create the bounding region of the shortest possible classification timeframe.
  We identified the period
  \DTMDate{2019-04-21} to \DTMDate{2019-08-09}
  as having the best trade-off in terms of accuracy and earliness.
  This timeframe only suffers a \qty{0.75}{\percent} loss in accuracy as compared to using the full timeseries.
  We observed that the \gls{gl:lrp}-derived important timesteps also highlight small details in input values that differentiates between different classes and possibly offers links to physical crop growth milestones.
\end{abstract}


\glsresetall

\section{Introduction}
\label{sec:intro}

Crop classification or the identification of crops grown on agricultural land is important for many administrative and planning purposes.
\Gls{gl:dl} algorithms provides efficient and highly accurate methods for crop classification \cite{kussul_2017}.
Various algorithms, such as convolution neural networks, recurrent neural networks, and transformer architectures, have achieved great success with accuracies ranging from \qtyrange[range-phrase={ to almost }]{79}{99}{\percent} \cite{russwurm_korner_2020, ji2018, kussul_2017, zhao_chen_jiang_jing_sun_feng_2019, li_2020}.
However, since ground truth data is only available \textit{post-hoc} after the growing season, established classification algorithms either consider year-long timeseries data or timeseries covering the growing season which requires expert knowledge.
There is, however, demand for early crop classification as that enables better water resource management and food security \cite{azar_villa_stroppiana_crema_boschetti_brivio_2016}.
Furthermore, with worsening effects of climate change, strategic planning involving crop map knowledge during the growing season itself becomes increasingly crucial.
The question arises whether shorter timeseries can be used as the input for the classification without sacrificing accuracy.

One way to answer this question uses trial-and-error strategies, such that the dataset is systematically tested over different time intervals to analyze the trade-off between earliness and accuracy \cite{azar_villa_stroppiana_crema_boschetti_brivio_2016, zhao_chen_jiang_jing_sun_feng_2019}.
These attempts determined the months of July and August being the most important for crop classification in the northern hemisphere, while rarely offering a definitive answer to the optimal timeframe \cite{zhao_chen_jiang_jing_sun_feng_2019,Inglada2016,cai2018}.
Other methods amend the cost functions directly with additional terms for earliness.
This balances the length of the timeseries with the achieved accuracy of classification when training the model~\cite{russwurm_korner_2019}.

We propose an alternative approach to isolate the shortest possible input data timeframe by identifying timesteps particularly important for performing that classification decision.
At its core, we employ \gls{gl:lrp}~\cite{bach_binder_montavon_klauschen_mueller_samek_2015,montavon_binder_lapuschkin_samek_mueller_2019}, an \gls{gl:xai} method to attribute to which extent an input affects the final classification result through calculating the \emph{relevance} of each layer from the desired output back to each particular input.
The higher the positive relevance score, the more influential that input is considered to be on the output, while a more negative relevance score discourages the output. %
That approach comes with the added benefit of connecting the model to real-life occurrences that better explains the model functionality and limitations.

\section{Data and Methodology}
We used the Austrian region of the EuroCrops~\cite{Schneider2021EPE} demo dataset TinyEuroCrops~\cite{Schneider2021TEC} that was collected based on the farmers' self declarations of the year 2019.
The dataset consists of \num{414392} parcels annotated with 44 crop classes.
Each parcel is represented by 13 spectral band values derived from Sentinel 2 images acquired between 
\timerange{2019-01-6}{2019-12-27}[{ through }].
The dataset is split into a train and a test region containing \numlist{345970;68422} parcels, respectively, that are spatially independent from each other.
Notably, our dataset is extremely imbalanced, where \qty{64}{\percent} of the parcels represent four crop classes while 29 crops classes are represented by less than \qty{1}{\percent} of the parcels each.

The methodology consists of three components.
First, to establish a baseline model, we trained an attention transformer model~\cite{garnot2020,Schneider21:SIT} using \qty{80}{\percent} of the parcels from the train region for crop classification, while using the remaining \qty{20}{\percent} for testing.
The transformer architecture consists of an initial \gls{gl:mlp} for feature extraction followed by a four-headed attention mechanism and two more \glspl{gl:mlp} as a decoder.

Secondly, to identify the key timesteps for classification, we carried out \gls{gl:lrp} analysis on the trained baseline model.
\Gls{gl:lrp} attributes the relevance score
\begin{align}
	{\relevance}^{(l)}_i &= \sum_{j} \frac{z_{ij}}{ \sum_{i^\prime}{z_{i^\prime j}}}{R^{(l+1)}_j} &
	\text{with}\; z_{ij} &= {x^{(l)}_i}{w^{(l,l+1)}_{ij}}
\end{align}
of one input neuron $i$ in layer $l$ by considering the fraction of contributions to
the outputs that comes from specific input neuron $i$  \cite{bach_binder_montavon_klauschen_mueller_samek_2015, montavon_binder_lapuschkin_samek_mueller_2019, pmlr-v162-ali22a}.
As \relevance is assigned to each input, which in the given dataset corresponds to each of the 13 spectral bands $b$ at every timestep $t$, we define the relevance of each timestep
\begin{equation}
	{{\relevance}_t} = \sum_{b} {\relevance^\text{input}_{b,t}} \;.
\end{equation}

Lastly, using the identified $R_t$, we derived the shortest timeframes based on the top $n$ largest absolute $R_t$.
We isolated the longest timerange $\Delta t_n$ for $n \in \{ 3, 5, 10\}$ bounded by the $n$ timesteps.

For testing, we pruned the timeseries in the test region according to the identified timeframes. These pruned timeseries and the complete timeseries from the test region are trained with the same attention transformer architecture during initial training.

\section{Results}
\subsection{Transformer Model and LRP Verification}
We successfully trained a crop classification model with \qty{85.76}{\percent} accuracy, which has comparable performance to existing \gls{gl:dl}-based models with large number of classes \cite{russwurm_korner_2020, Schneider21:SIT}.
With confidence in our trained model, we applied \gls{gl:lrp} analysis.
We further confirmed that \gls{gl:lrp} is capable of attributing relative importance to each timestep by comparing to a random baseline in a pruning experiment \cite{pmlr-v162-ali22a}.
We carried out pruning of the input timeseries- either targeted based on the least importance as identified by \gls{gl:lrp} or randomly and fed the successively pruned timeseries into the trained attention model to calculate the mean squared error from the full timeseries output logits.
The results confirmed our hypothesis and showed that targeted pruning lead to a slow initial increase in error followed by exponential increase in error towards the end.
In contrast, random pruning lead to a linear increase in error.

\subsection{LRP-Identified Timesteps}
\begin{figure}[t]
  \subcaptionbox{\gls{gl:sunflower} input}
                {\includegraphics[width=.5\linewidth]{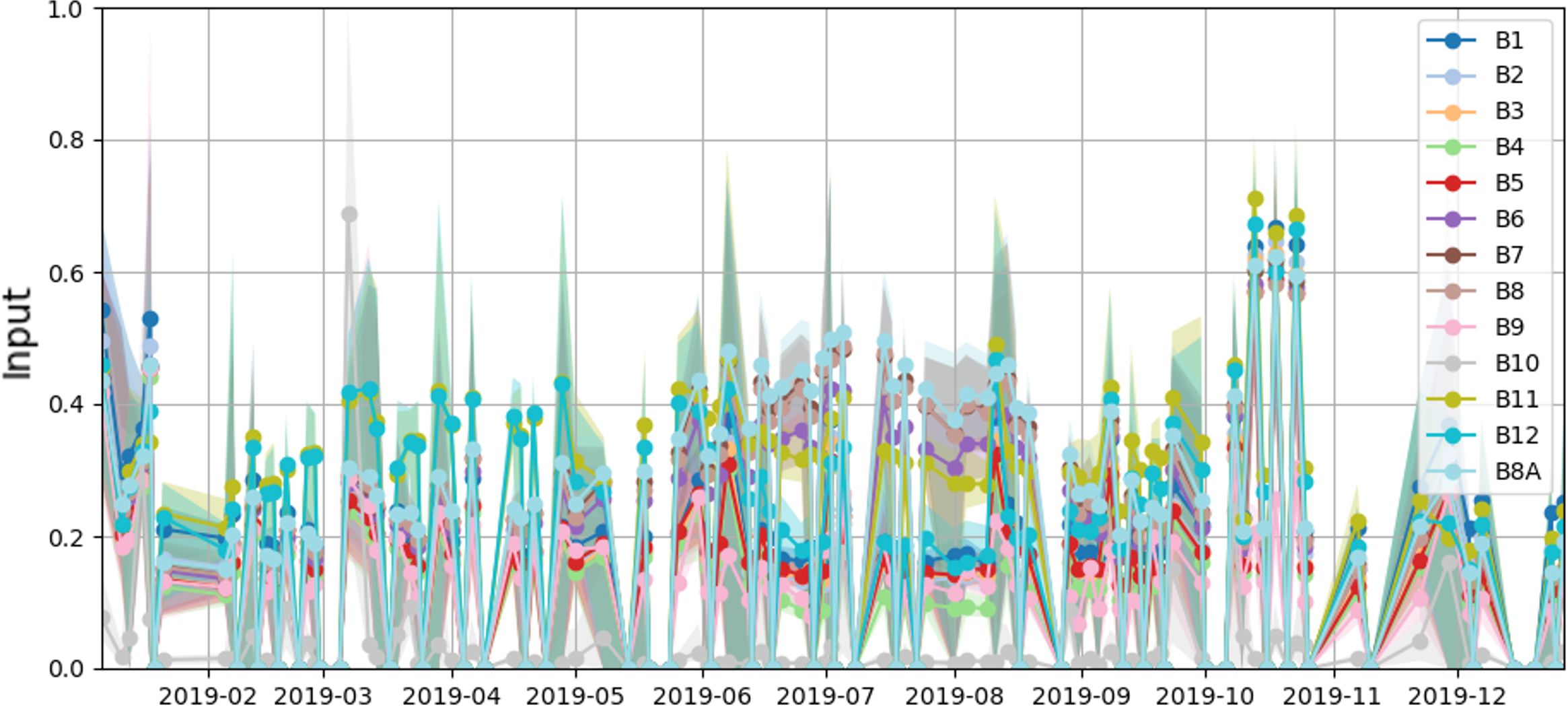}}%
  \subcaptionbox{\gls{gl:sunflower} \relevance}
                {\includegraphics[width=.5\linewidth]{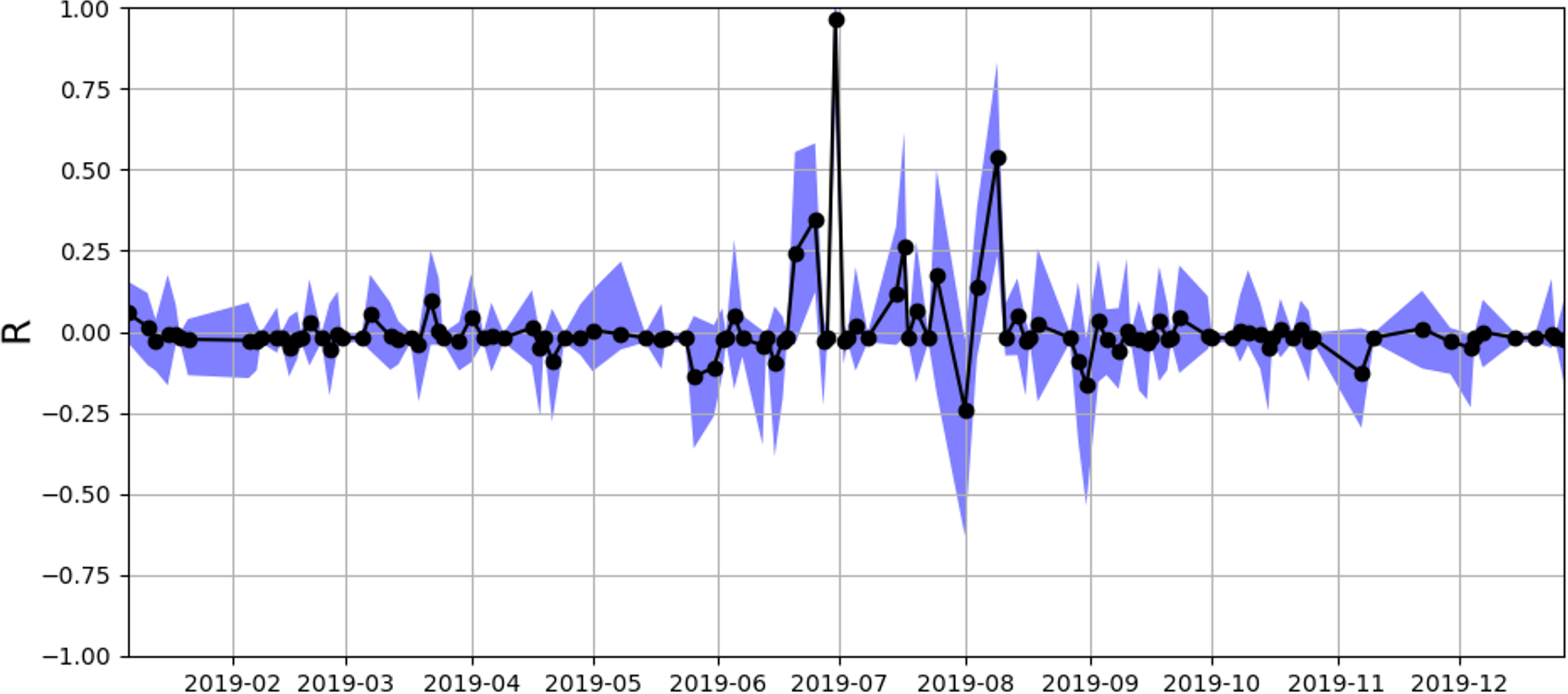}}
  \subcaptionbox{\class{Cucurbits} input}
                {\includegraphics[width=.5\linewidth]{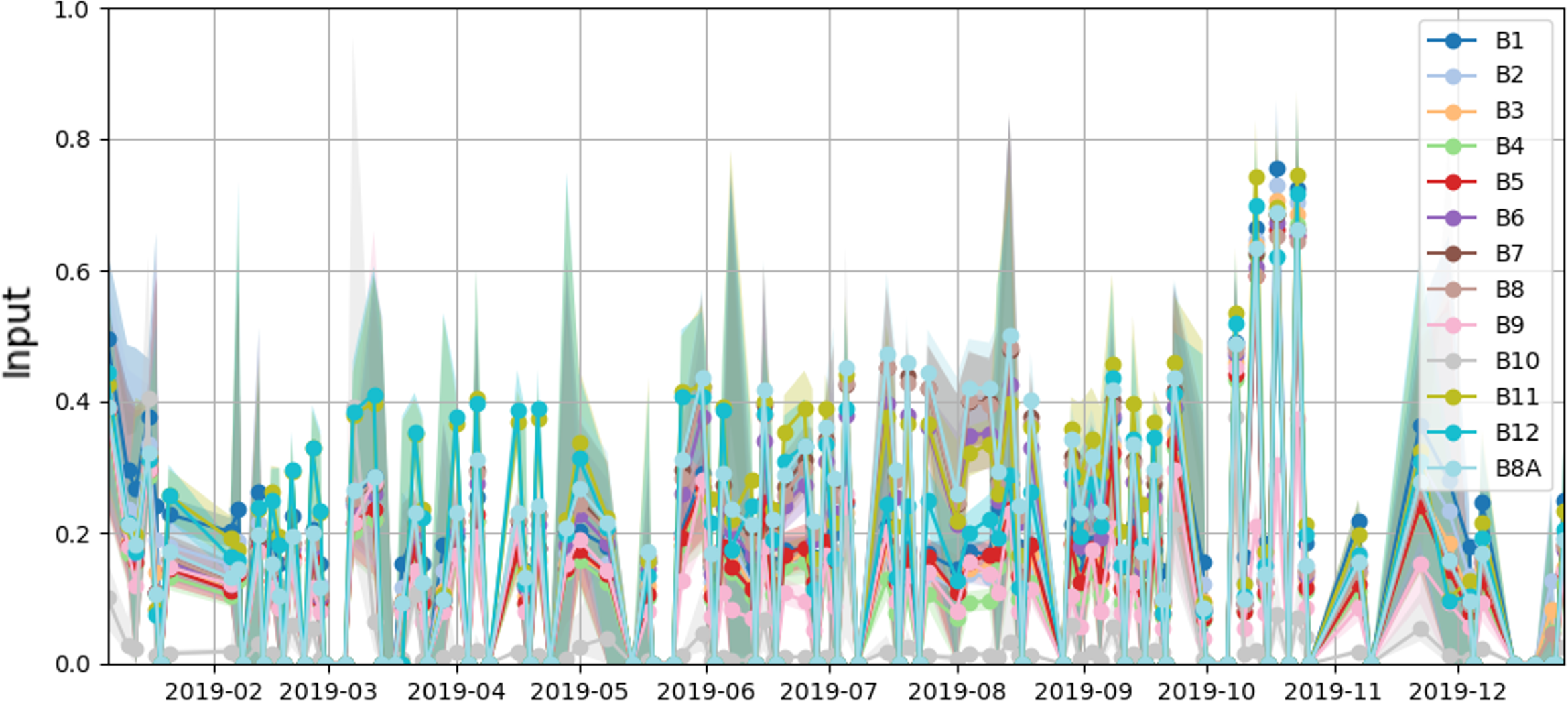}}%
  \subcaptionbox{\class{Cucurbits} \relevance}
                {\includegraphics[width=.5\linewidth]{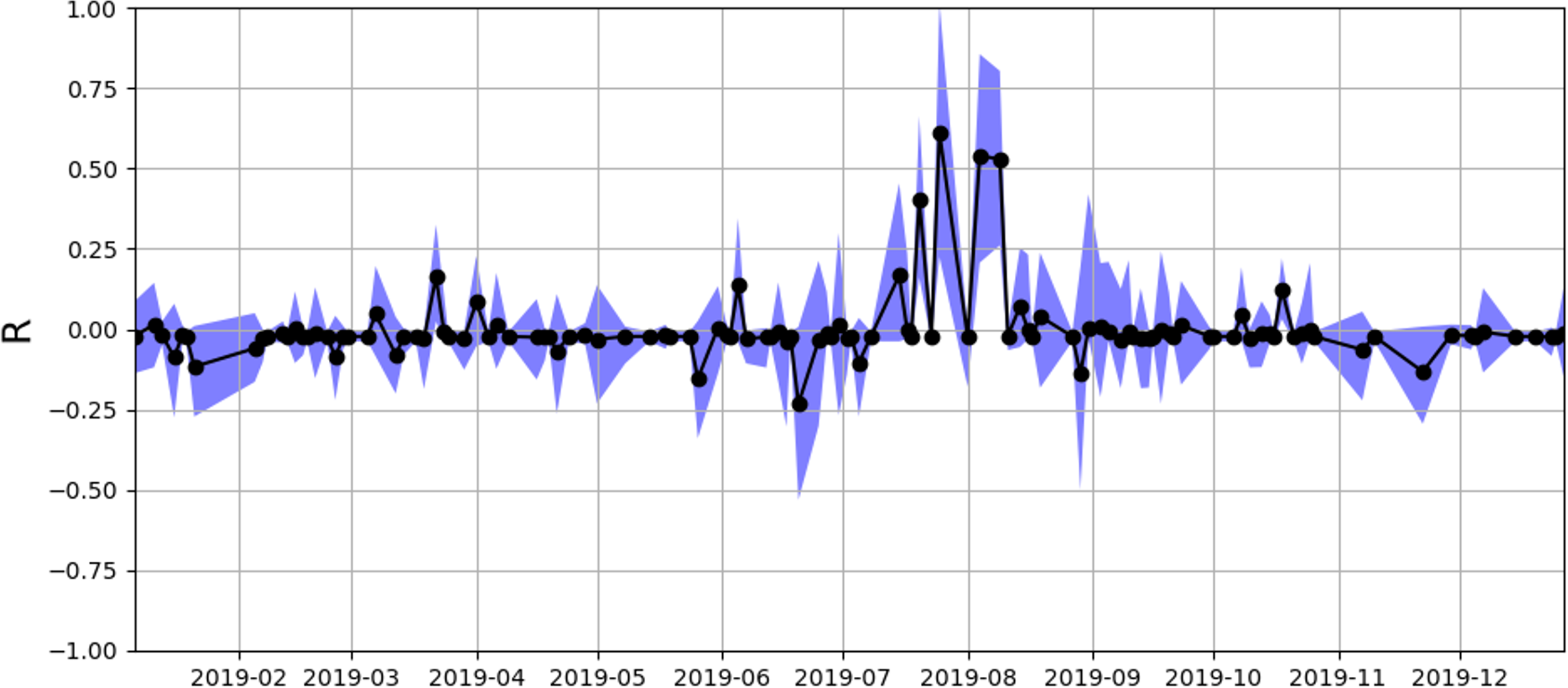}}
  \subcaptionbox{\gls{gl:corn} input}
                {\includegraphics[width=.5\linewidth]{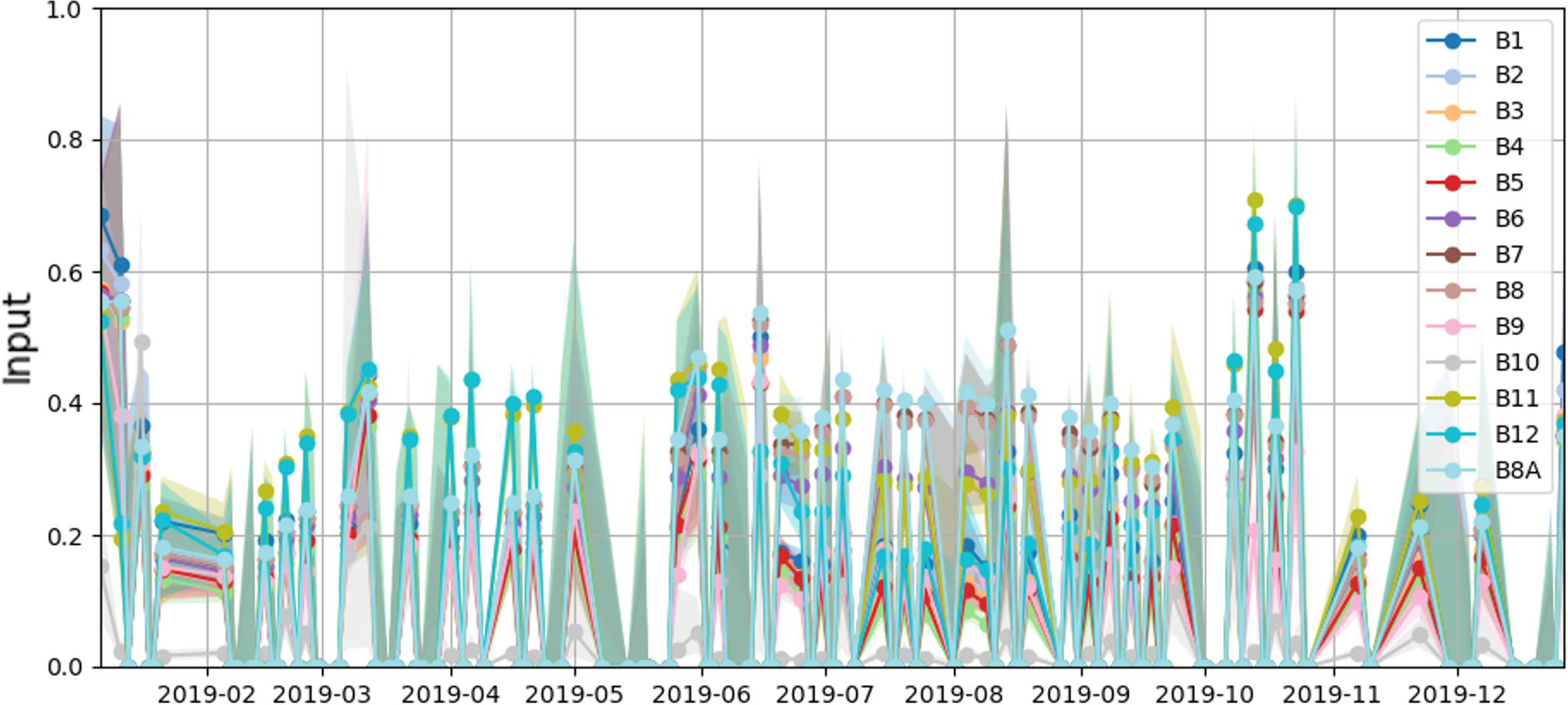}}%
  \subcaptionbox{\gls{gl:corn} \relevance}
                {\includegraphics[width=.5\linewidth]{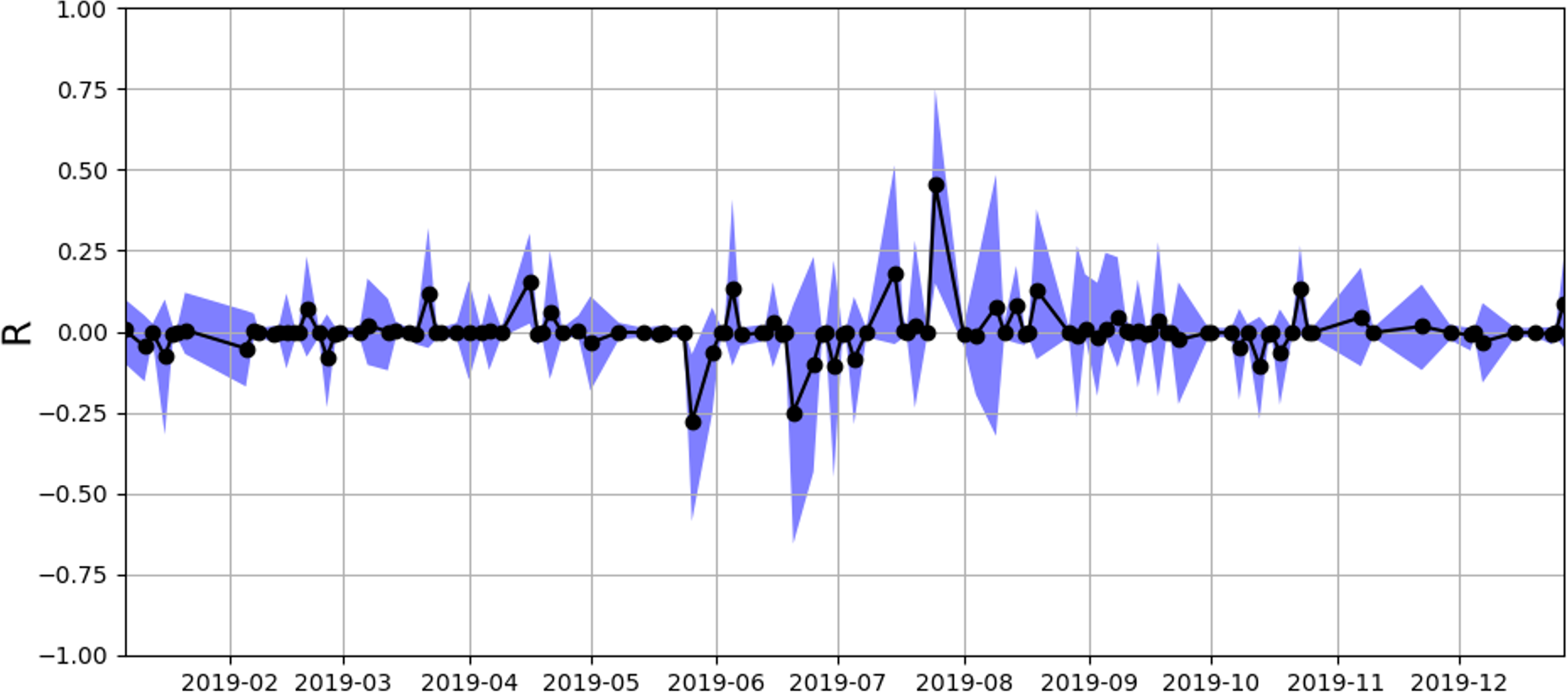}}
  \caption{%
    Input spectral reflectances and corresponding $R_t$ for three summer crops.
    The solid black line represents the median $R_t$ for the crop type while the shaded blue region represents the 25 to 75 percentile.
    For each timeseries only a few dominant peaks that have significantly larger $R_t$ exist.
    These peaks tend to be crop-specific and represents the most important timesteps for the classification of that crop. For \Gls{gl:sunflower}, this is in late June and early August,  \class{Cucurbits} have them in mid July to early August and \Gls{gl:corn} have them in July.
  }
  \label{fig:rvals}
\end{figure}

Overall, only a few timesteps in a year showed strong positive \relevance values, as visualized in \cref{fig:rvals}.
These timesteps are the most important to classification and \relevance peaks at different times depending on the crop.
We demonstrate the effects of \gls{gl:lrp} using three of the best performing crop classes for both producer's and user's accuracy.

These are the typical summer crops \gls{gl:sunflower}, \class{Cucurbits}, and \gls{gl:corn}.
Despite having a similar growing season, while \glspl{gl:sunflower} have important timesteps in late June and early August, \class{Cucurbits} have them in mid July to early August and \gls{gl:corn} have them in July.

\begin{figure}[t]%
	\centering
	\includegraphics[width=\linewidth]{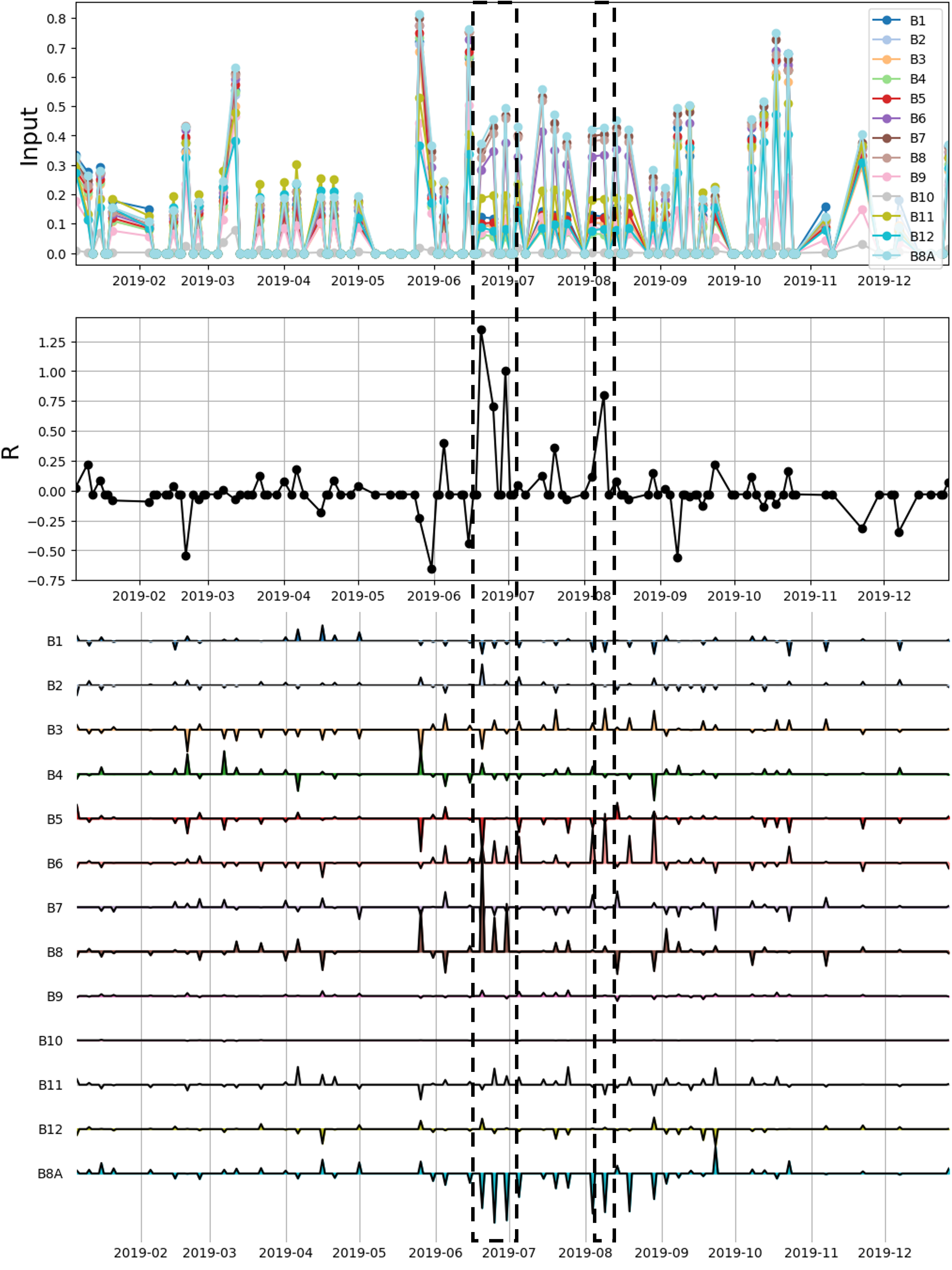}
	\caption{
		A sample \gls{gl:sunflower} parcel with the input values at the top, the relevance scores at each timestep $R_t$ in the middle and the corresponding band-specific relevance scores at each timestep $R_{b,t}$ in the bottom.
    The four dominant peaks indicate the most important timesteps are from mid to late June and early August.
    The corresponding important bands are B6, B8 and B8A}
	\label{fig:case}
\end{figure}

When compared to the input spectral reflectances, we clearly observe a difference in the input values in late June for \gls{gl:sunflower} as compared to the other two classes.
For \gls{gl:sunflower}, reflectances in band B6-B8 and B8A recorded higher values than B11 which is the spectral band with the highest value in \gls{gl:corn} and \class{Cucurbits}.
Detailed analysis of ${R_{b,t}}$ for \gls{gl:sunflower} using a typical parcel in \cref{fig:case} revealed high ${R_{b,t}}$ values for B6 and B8.

\Gls{gl:corn} and \class{Cucurbits} have relatively similar distributions in ${R_t}$ and input.
Most noticeably different input values occur during timesteps with cloud coverage or missing values.
These cloud covered parcels typically have high values in all bands.
This is the case in mid June for \Gls{gl:corn} where ${R_t}$ is close to 0.
The high ${R_t}$ timesteps in \class{Cucurbits} from mid July to early August revealed a minute detail in the input timeseries of \class{Cucurbits}.
Only during this period, the reflectances in some bands like B6 are higher as compared to \Gls{gl:corn}.
Detailed band-specific breakdown of the relevance scores revealed high ${R_{b,t}}$ values assigned to B6-B8 for \class{Cucurbits} at these timesteps.

Past studies on \gls{gl:corn} identified the silking stage as an important time point for classification \cite{xu_yang_xiong_li_huang_ting_ying_lin_2021}.
This corresponds to mid July in Austria, which has been identified as an important timestep \cite{eitzinger2013}.

\subsection{Early Crop Classification Timeframes}
The results summarized in \cref{tbl:results} showed close to no degradation in accuracy of the pruned datasets.
The identified timeframes are
\timerange{2019-04-21}{2019-08-09}
when using three important timesteps ($\Delta t_3$),
\timerange{2019-04-01}{2019-09-18}
when using five important timesteps ($\Delta t_5$), and
\timerange{2019-02-20}{2019-10-21}
when using ten important timesteps ($\Delta t_{10}$).

\begin{table}[t]
  \centering
  \caption{%
    Training and Testing accuracy (in \unit{\percent}) of timeframes selected using different number of important timesteps.\newline
    Full data: \timerange[abbr]{2019-01-06}{2019-12-27},
    $\Delta t_3$: \timerange[abbr]{2019-04-21}{2019-08-09},
    $\Delta t_{5}$: \timerange[abbr]{2019-04-01}{2019-09-18},
    $\Delta t_{10}$: \timerange[abbr]{2019-02-20}{2019-10-21}}
  \label{tbl:results}

  \begin{tabular}{ccccc}
    \hline
    & Full Data & $\Delta t_3$  & $\Delta t_{5}$  & $\Delta t_{10}$  \\
    \hline
    Train & 85.76 & --- & --- & ---    \\
    Test  & 87.15 & 86.40 & 86.96  & 87.09    \\
    \hline
  \end{tabular}
\end{table}

In our experiments, $\Delta t_3$ offered the best trade-off in terms of accuracy and earliness, as it ends almost two months before the end of the growing season while compromising less than \qty{1}{\percent} in accuracy as compared to using the full dataset.
Interestingly,  $\Delta t_{10}$ covers dates outside the growing season.
We suggest only few timesteps play a significant role in classification and by including 10, the peaks with much smaller \relevance values are also included.
For example, in \cref{fig:rvals} the number of significant absolute $R_t$ peaks (larger than 0.25) are: three for \gls{gl:sunflower}, four for \class{Cucurbits}, and two for \gls{gl:corn}.
We call these peaks \emph{dominant peaks} and argue those without clear dominant peaks lack any significant timesteps for classification.
For example, the class \class{Fallow Land Not Crop} only found relevance scores $R \in \interval{-0.25}{0.25}$ and, correspondingly, observed one of the lowest producer accuracies of \qty{67.51}{\percent}.

Our identified timeframes offer comparable results to similar studies on early crop classification.
Early crop classification in neighboring Germany requires until mid August for \qty{75}{\percent} classification accuracy of all seven crops \cite{russwurm_korner_2019}.
In Midwestern United States, despite finding a 5 week window between late July to late August with a small \qty{0.2}{\percent} loss in accuracy, the other tested timeframes had accuracies decrease by \qtyrange{15}{30}{\percent} \cite{xu_yang_xiong_li_huang_ting_ying_lin_2021}, which suggest our loss in accuracy of \qty{0.75}{\percent} can be considered negligibly small.

\section{Conclusion and Future Work}

This early study showed a simple and effective method for isolating the important timeframe for crop classification. \Acrlong{gl:lrp} was able to successfully identify meaningful inputs that differentiates between the different crop classes and highlight the most important timesteps to classifications.
The identified timesteps is potentially connected growth milestones in real life, as in the case of silking stage of \gls{gl:corn}.

Future work will focus on transferring these relevant timeframes across different years to consider the effects of interannual variability so that a climatological timeframe for classification can be reliably obtained.

\section{References}
\label{sec:ref}

\setlength{\bibitemsep}{1pt}
\defbibenvironment{bibliography}
{\list
  {\printtext[labelnumberwidth]{%
      \printfield{labelprefix}%
      \printfield{labelnumber}}}
  {\setlength{\labelwidth}{\labelnumberwidth}%
    \setlength{\leftmargin}{\labelwidth}%
    \setlength{\labelsep}{\biblabelsep}%
    \addtolength{\leftmargin}{\leftskip}%
    \setlength{\itemsep}{\bibitemsep}%
    \setlength{\parsep}{\bibparsep}}%
  \renewcommand*{\makelabel}[1]{\hss##1}}
{\endlist}
{\item}

\renewcommand{\UrlFont}{\ttfamily\scriptsize}
\printbibliography[heading=none]

\end{document}